\documentclass[twocolumn]{article}

\usepackage[a4paper, textwidth=7.2in, top=1in, bottom=1in]{geometry}

\usepackage[utf8]{inputenc}
\usepackage[round]{natbib}
\usepackage[colorlinks=true, allcolors=blue, hypertexnames=false]{hyperref}
\usepackage[font=small, labelfont=bf]{caption}

\usepackage{mathpazo}    
\usepackage{graphicx}    
\usepackage{amsmath}     
\usepackage{amsfonts}    
\usepackage{booktabs}    
\usepackage{microtype}   
\usepackage[misc]{ifsym} 

\usepackage{multirow}    
\usepackage{makecell}    

\usepackage{titling}
\pretitle{\begin{flushleft}\rule{\textwidth}{0.8pt}\end{flushleft} \begin{center}\huge\sc}
\posttitle{\par\end{center} \rule{\textwidth}{0.8pt} \vskip 0.5em}

\usepackage{authblk}

\usepackage{orcidlink}

\title{Most biomedical publications show signs of LLM-assisted writing}

\author[1]{Lena Holzwarth\,\orcidlink{0009-0005-5968-0113}\,}
\author[1]{Rita González-Márquez\,\orcidlink{0009-0005-6840-7979}\,}
\author[1,2,3]{Dmitry Kobak\,\orcidlink{0000-0002-5639-7209}\,}    

\affil[1]{Hertie Institute for AI in Brain Health, University of T\"ubingen, Germany}
\affil[2]{Department of Mathematics, Computer Science, and Statistics, Ghent University, Belgium}
\affil[3]{VIB Center for AI and Computational Biology, VIB, Ghent, Belgium}

\affil[ ]{}
\affil[ ]{\Letter\: \normalfont{\texttt{dmitry.kobak@ugent.be}}}

\usepackage{fancyhdr}
\usepackage{abstract}
\begin{document}

\twocolumn[
\begin{@twocolumnfalse}

\maketitle

\begin{abstract}
Over the past several years, LLM-powered chatbots and agents have become widely used as a tool for academic writing. LLM-assisted writing can be valuable by removing language barriers but at the same time causes concerns about misconduct and fraud. To inform policy decisions, it is necessary to monitor the prevalence of LLM-altered texts in scholarly publications. Despite some recent progress in this direction, no existing method can produce reliable estimates. Here we suggest and validate a new unbiased approach to estimate LLM usage in a corpus of texts based on changing word frequencies. We apply our method to the full texts of open-access biomedical papers from Pubmed Central, and show that by the end of 2025, 89\% of papers show excess of LLM-associated vocabulary. We also find that LLMs are twice as likely to be used when writing a paragraph in the Discussion section (68\%) compared to a paragraph in the Methods section (32\%), but even inside the Methods section, the overall prevalence of LLM usage is over 50\%. We believe that our estimates are crucial to shape future guidelines and policies. 
\end{abstract}

\vspace{2em}
\end{@twocolumnfalse}
]

\section{Introduction}

\begin{figure*}[t]
    \centering
    \includegraphics[width=\textwidth]{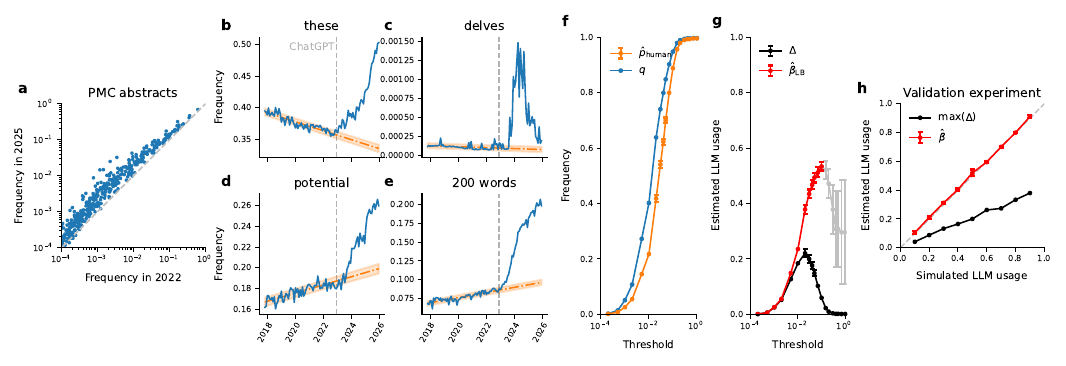}
    \caption{\textbf{Overview of the estimation and validation process.} \textbf{(a)} Usage frequencies of 379 LLM-associated marker words in PMC abstracts before (2022) and after (2025) widespread LLM adoption. \textbf{(b--d)} Observed ($q$) and expected ($\hat{p}_{\text{human}}$) frequencies for individual words in PMC abstracts. Linear regression is fit to the 60 months in 2018--2022. \textbf{(e)} The same for a set of 200 rarest marker words. \textbf{(f)} Observed and expected frequencies in 2025 PMC abstracts for different sets of marker words, each containing all marker words with frequency below a given threshold. Here $\hat{p}_{\text{human}}$ is computed based on yearly data with regression based on five years 2018--2022. \textbf{(g)} Lower bounds on the LLM usage derived from (f). Thresholds yielding standard error of $\hat\beta_\text{LB}$ above 0.025 are shown in gray and discarded. The maximum remaining value of $\hat{\beta}_{\text{LB}}$ is chosen as the LLM usage estimate $\hat{\beta}$. \textbf{(h)} Validation experiment with ground-truth values of LLM usage $\beta$ correctly retrieved by $\hat{\beta}$, and strongly underestimated by the maximal frequency gap $\max(\Delta)$.}
    \label{fig:method}
\end{figure*}

Since the release of OpenAI's ChatGPT in November 2022, chatbots and agents based on large language models (LLMs) have become increasingly helpful in any workplace dealing with writing, such as academia. While LLMs can increase inclusivity and accessibility for non-native English speakers, there are concerns about hallucinations \citep{Walters2023, topaz2026fabricated, zhao2026llm} and fraud \citep{Kendall2024} that can negatively impact academic publishing. To inform any policy decisions, it is important to estimate the prevalence of actual LLM usage.

While LLMs produce text that can be difficult to distinguish from human-written, they introduce effects visible on the corpus level. For example, certain style words appear with higher frequency in LLM-generated texts, affecting word frequencies in a corpus of documents, as observed in publications from, e.g., linguistics \citep{Botes2025}, astronomy \citep{Astarita2024}, medicine \citep{Matsui2024}, and dentistry \citep{Uribe2024}. Comparing the pre-LLM and post-LLM frequencies of LLM-associated marker words, either pre-selected \citep{gray2024chatgpt, Gray2025, Kousha2025}, or optimized  \citep{kobak2025delving, siler2026diffusion} can yield an estimate of the overall LLM usage. However, all existing methods in this group can only provide a lower bound on the LLM usage. Another approach is to measure word frequencies in LLM-generated texts and then employ a mixture model to estimate LLM usage in a corpus \citep{liang2024monitoring, Liang2025, Geng2024}. The inherent limitation here is that the results can strongly depend on the exact prompts and models used to generate LLM texts. 

In this work, we go beyond prior literature in two respects. First, we suggest and validate a simple method to estimate LLM usage in an assumption-free way based on a large set of marker words, going beyond the lower bounds. Second, we apply this method to analyze full texts of open-access biomedical papers from PubMed Central (PMC), which allows us to compare LLM usage across different sections of the papers, such as Methods, Results, and Discussion. Our estimate of the prevalence of LLM-assisted writing in PMC in the end of 2025 is 89\%, which is higher than previously reported estimates.

\section{Results}

We based our analysis on PubMed Central (PMC), a digital repository of open-access scholarly articles published in biomedical and life-science journals. We accessed PMC in early 2026 and restricted our analysis to 1,194,287 papers written in English, published in 2017--2025, and containing sections called Introduction, Methods, Results, and Discussion (or equivalent names; see Methods).

In our prior work \citep{kobak2025delving}, we identified 379 non-content words that showed markedly increased usage in PubMed abstracts in 2024 compared to earlier years, and attributed this excess to LLM-assisted writing or editing. These LLM markers are words that can be used in any research context, such as \textit{these}, \textit{potential}, or \textit{delves}. We confirmed that in our PMC dataset almost all of these words saw increased usage in abstracts in 2025 (Figure~\ref{fig:method}a), so we based our subsequent analysis on this list.

For any single of the marker words, we obtained its usage frequency $q(t)$, defined as the fraction of documents containing at least one occurrence of the marker word, as a function of time (Figure~\ref{fig:method}b--d). We fit a linear regression to the five-year period (2018--2022) before the widespread adoption of LLMs (ChatGPT was released in November 2022), and extrapolated the regression line to 2023--2025 to obtain $\hat{p}_\text{human}(t)$, the counterfactual usage frequency of this word in human-written texts in the absence of LLMs. The frequency gap $\Delta=q-\hat{p}_\text{human}$ gives a lower bound on LLM usage \citep{kobak2025delving}: for example, for the word \textit{these} in abstracts in December 2025, we have $q=0.50$ and $\hat{p}_\text{human}=0.33$ (Figure~\ref{fig:method}b), meaning that the excess 17\% usage must be due to LLM editing.

A tighter lower bound is also possible. If $p_\text{LLM}$ is the (unobservable) usage frequency in LLM-edited or LLM-written texts, and $\beta$ is the proportion of papers written with some help of an LLM, then 
$$q = (1-\beta) p_\text{human} + \beta p_\text{LLM}.$$
From here we obtain
$$\beta = \frac{q-p_\text{human}}{p_\text{LLM}-p_\text{human}} \geq \frac{q-p_\text{human}}{1-p_\text{human}} \geq q-p_\text{human}.$$
To calculate $\beta$, we would need access to $p_\text{human}$ and $p_\text{LLM}$. 
For the former, we can plug in our estimate $\hat p_\text{human}$, but the latter cannot be inferred directly from the data. 
By replacing it with its upper bound $p_\text{LLM}\leq1$, we arrive at a lower bound that is stronger than $\Delta$:
$$\beta \ge \hat \beta_\text{LB} = \frac{q - \hat p_\text{human}}{1 - \hat p_\text{human}} \geq \Delta.$$
For the same word \textit{these} in PMC abstracts in December 2025, we get $\hat\beta_\text{LB}=0.17/0.67=0.25$, implying at least 25\% LLM usage.

\begin{table}[t]
\centering
\begin{tabular}{lcccc}
    \toprule
    & 2023 & 2024 & 2025 & Dec 2025 \\
    \midrule
    & \multicolumn{4}{c}{\textit{Entire sections}} \\
    Abstract                   & .09 \scriptsize{(2)} & .31 \scriptsize{(2)} & .53 \scriptsize{(1)} & .68 \scriptsize{(2)} \\
    Introduction               & .11 \scriptsize{(2)} & .34 \scriptsize{(2)} & .50 \scriptsize{(2)} & .63 \scriptsize{(2)} \\
    Methods                    & .11 \scriptsize{(2)} & .29 \scriptsize{(2)} & .46 \scriptsize{(2)} & .54 \scriptsize{(2)} \\
    Results                    & .11 \scriptsize{(3)} & .30 \scriptsize{(3)} & .47 \scriptsize{(2)} & .58 \scriptsize{(2)} \\
    Discussion                 & .14 \scriptsize{(2)} & .43 \scriptsize{(2)} & .68 \scriptsize{(1)} & .78 \scriptsize{(2)} \\
    \textbf{Full paper}        & \textbf{.19 \scriptsize{(5)}} & \textbf{.52 \scriptsize{(4)}} & \textbf{.77 \scriptsize{(2)}} & \textbf{.89 \scriptsize{(3)}} \\
    \midrule
    & \multicolumn{4}{c}{\textit{Random 255-word crops}} \\
    Abstract                   & .10 \scriptsize{(2)} & .32 \scriptsize{(2)} & .54 \scriptsize{(2)} & .67 \scriptsize{(2)} \\
    Introduction               & .10 \scriptsize{(2)} & .30 \scriptsize{(1)} & .48 \scriptsize{(1)} & .59 \scriptsize{(2)} \\
    Methods                    & .03 \scriptsize{(2)} & .14 \scriptsize{(2)} & .28 \scriptsize{(2)} & .32 \scriptsize{(4)} \\
    Results                    & .08 \scriptsize{(1)} & .20 \scriptsize{(1)} & .35 \scriptsize{(1)} & .46 \scriptsize{(2)} \\
    \textbf{Discussion}        & \textbf{.10 \scriptsize{(2)}} & \textbf{.35 \scriptsize{(2)}} & \textbf{.57 \scriptsize{(2)}} & \textbf{.68 \scriptsize{(2)}} \\
    Full paper                 & .07 \scriptsize{(2)} & .21 \scriptsize{(2)} & .36 \scriptsize{(1)} & .40 \scriptsize{(3)} \\
    \bottomrule
\end{tabular}
\caption{Estimated frequencies of LLM usage for each manuscript section. Each row used its own threshold $T$, which was optimized for 2025 (and then used for all years). Top part: entire sections. Bottom part: random 255-word crops. The row with the highest values in each part is highlighted in bold. `Full paper' refers to Introduction, Methods, Results, and Discussion concatenated together. Standard errors for the last digit are shown in parentheses.}
\label{tab:results}
\end{table}

As each of the marker words is associated with LLM usage, using a set of marker words $G$ to compute the fraction of documents $q(t)$ containing at least one occurrence of at least one of the words from $G$ can give higher bounds on $\beta$. However, making $G$ too large can lead to $\hat p_\text{human} \approx 1$, leading to unstable estimates with large standard errors (see Methods), so some selection has to be made. Following \citet{kobak2025delving}, we sorted all marker words by their frequency in 2024, and selected all words with $q<T$ into a set $G(T)$, parameterized by the threshold $T$. As an example, selecting $T$ to yield 200 words in PMC abstracts (Figure~\ref{fig:method}e), resulted in $\hat \beta_\text{LB} = 0.11$ in December 2025.

To find an optimal value of $T$, we used a grid of $T$ values. For each $T$ value, we found $q$ and $\hat{p}_\text{human}$ for 2025 using regression on yearly 2018--2022 frequencies (Figure~\ref{fig:method}f), and selected $T$ with the highest $\hat \beta_\text{LB}$ (Figure~\ref{fig:method}g), after discarding all $T$ values yielding standard error above $0.025$ (see Methods). As this maximum was typically achieved at $q$ close to 1 (and hence $p_\text{LLM}$ close to 1, which is the approximation used in computing the lower bound), we assume that our lower bound is acceptably tight and use it as the estimate of LLM usage frequency $$\hat\beta=\max_T(\hat\beta_\text{LB}).$$

We used a simulation to validate our estimation procedure and could retrieve the ground-truth $\beta$ with high accuracy (see Methods for details). The simulation assumed 100,000 documents per year, which was close to the amount of papers in our PMC sample, and realistic ranges for $p_\text{human}$ and $p_\text{LLM}$ (see Methods). We obtained $|\hat \beta - \beta| < 0.02$ for all simulated values $\beta\in[0,1]$ (Figure~\ref{fig:method}h), confirming the reliability of our estimation procedure.

\begin{figure}[t]
    \centering
    \includegraphics[width=\linewidth]{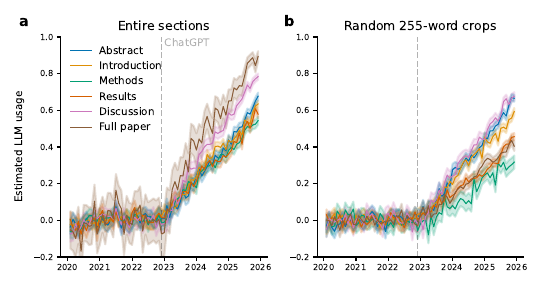}
    \caption{\textbf{Estimated LLM usage for individual sections}. \textbf{(a)} Estimated LLM usage over time using entire sections. All estimates were computed with the section-specific optimal threshold for the year 2025. The standard errors are shown as shaded areas. \textbf{(b)} The same for random 255-word crops from each section.}
    \label{fig:curves}
\end{figure}

We applied this procedure to estimate the frequency of LLM usage with yearly (Table~\ref{tab:results}) and monthly (Figure~\ref{fig:curves}a) resolution for each of the PMC manuscript sections. Threshold $T$ was optimized for each section separately using the 2025 data, and then used for other years and months as well. All sections showed continually increasing estimated LLM usage starting in early 2023. For the PMC abstracts, we obtained $\hat\beta = 0.53$ in all 2025 and 0.68 in December 2025. For the full paper (Introduction, Methods, Results, and Discussion concatenated together), we obtained $\hat\beta = 0.77$ in all 2025 and 0.89 in December 2025. The interpretation is that by the end of 2025, 9 out of 10 papers exhibited signs of LLM-assisted writing or editing.

As expected, the estimates for the individual sections were all lower than the estimate for the full paper text, as our approach detects LLM usage in any part of the paper. Similarly, comparing different sections with each other is confounded by their average length, because a long section presents more opportunity for authors to employ LLM editing and for an LLM to use its marker words. Therefore, we repeated our estimation procedure using random 255-word crops from each section (corresponding to the median abstract length). Here, the Discussion had the highest $\hat\beta$ values in each year, reaching 0.68 in December 2025, closely followed by the abstracts reaching 0.67 (Table~\ref{tab:results}). We obtained the lowest LLM usage in the Methods section, reaching only 0.32 in December 2025. The Methods also had the biggest gap between the estimate based on the entire and on the cropped sections (0.54 vs. 0.32), indicating that LLM usage was not uniform over the length of this section.

\begin{figure}[t]
    \centering
    \includegraphics[width=\linewidth]{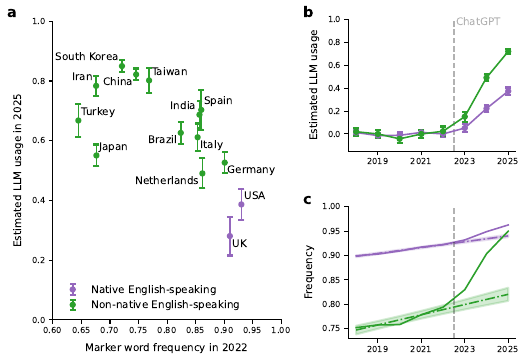}
    \caption{\textbf{Estimated LLM usage for individual countries (full papers).} \textbf{(a)} Estimated LLM usage in 2025 for some of the top 20 countries by paper count, plotted against the pre-LLM frequency of the set of selected marker words. Estimates with a standard error $\geq 0.07$ are not shown here but can be found in Table \ref{tab:countries}. The same set of 251 marker words was used for all countries (see Methods), so some of the values can be an underestimation. Countries are colored based on whether they have a majority of native English speakers. \textbf{(b--c)} Estimated LLM usage and marker word frequencies for all countries in PMC with a majority of native English speakers, and all other countries. Dashed lines show the expected frequencies.}
    \label{fig:countries}
\end{figure}

Finally, we sorted all PMC papers in our sample by country of affiliation of the first author and estimated LLM usage for each of the top 20 countries in terms of the number of papers in our dataset. We found the highest $\hat\beta$ in 2025 in South Korea (0.85), followed by China (0.82) and Taiwan (0.80), while the lowest value was observed in the UK (0.28), in agreement with prior studies reporting large differences in estimated LLM usage between affiliation countries \citep{Liao2024, Lin2025_NES, Prakash2025, kobak2025delving, siler2026diffusion}. Note that in this analysis we used a fixed set of marker words and did not optimize it for individual countries, so some of our estimates here can be an underestimation. The estimated LLM usage was negatively correlated with the frequency of LLM marker words before LLMs, i.e. $q(2022)$ (Figure~\ref{fig:countries}a), which can be taken as a rough proxy for the mastery of English language, as many marker words are uncommon vocabulary (e.g. \textit{exacerbating} or \textit{invaluable}). When grouping together all papers from countries with predominantly native English speakers (Australia, Canada, Ireland, New Zealand, South Africa, UK, USA) and the rest of the world, we obtained $\hat\beta$ values in 2025 of 0.37 and 0.72 respectively (Figure~\ref{fig:countries}b). The pre-2023 frequency of LLM marker words was much lower in non-natively English-speaking countries, but it became very close to the frequency in natively English-speaking countries by 2025 (Figure~\ref{fig:countries}c), indicating strong linguistic convergence driven by the LLM usage \citep{Lin2025_NES, Prakash2025}.

\section{Discussion}

\paragraph{Comparison to existing estimates}

We found that by the end of December 2025, almost 90\% of PMC papers showed signs of LLM-assisted writing or editing. Our estimates for 2023--2025 are higher than most existing estimates reported in the literature (Table~\ref{tab:review_estimation}), but are consistent with the usage ranges in existing surveys (Table~\ref{tab:review_surveys}).

Studies relying on word frequency gaps \citep{kobak2025delving, Gray2025, Kousha2025} reported estimates for 2024 ranging from 12\% to 16\% based on abstracts and full texts of papers from PubMed, PMC, and Dimensions databases, which is much smaller than our 2024 estimate of 31\% (abstracts) and 52\% (full papers). As we showed, this is because frequency gaps systematically underestimate the true fraction of LLM-assisted articles. A recent study based on the distribution of marker-word usage rates per 1000 words \citep{siler2026diffusion} found 57\% in 2025. This is also an underestimation, because this method is based on the overlap of the usage rate distributions pre- and post-2022 and implicitly assumes non-overlapping rates for human and LLM writing, which in reality is not the case. 

Studies based on measuring $p_\text{LLM}$ via prompting popular LLMs \citep{Geng2024, Liang2024_paper} produced very different results (35\% and 18\% respectively) on the same data (computer-science arXiv abstracts in early 2024), illustrating strong dependence on chosen prompts and models \citep{Dou2024}.
Existing studies based on LLM detectors \citep{Akram2024, Liu2024_profiles, Picazo-Sanchez2024} estimated LLM usage in 2023 when it was much lower.
Furthermore, various LLM detectors have been repeatedly found to be inaccurate \citep{Weber-Wulff2023, Lazebnik2024}.

Several surveys of scientists' use of LLMs for academic writing have been conducted by publishing houses \citep{Elsevier2024, Kwon2025, Nordling2023, VanNoorden2023, Wiley2025, Wiley2026} and independent researchers \citep{Eppler2024, Liao2024, Mishra2024, Mohammadi2026}. Many of them reported much higher usage (Table~\ref{tab:review_surveys}) than the estimation efforts reviewed above (Table~\ref{tab:review_estimation}). For example, \citet{Wiley2026} found over 70\% of survey respondents in mid-2025 using LLM for writing assistance. This is in good agreement with our estimates, especially taking into the account a possible underreporting bias in surveys. People are likely using LLMs more often than they are admitting when surveyed. Also, surveys typically ask about specific use cases, so actual LLM usage can be higher.

\paragraph{Limitations}

Our estimate hinges on the central assumption that $\hat{p}_\text{human}$ is a faithful estimate of $p_\text{human}$ in 2025. While linear extrapolation remains reasonable for some time after the release of ChatGPT, it becomes more tenuous as we move away from this date. As people become more and more exposed to the writing style of LLMs, they may start consciously or subconsciously copying it, and indeed there is some evidence for that in spoken communication \citep{Yakura2025}. On the other hand, there is also an opposite effect of people consciously avoiding LLM-associated marker words \citep{Geng2025} (Figure~\ref{fig:method}d), and it is unclear which effect is stronger. We believe that any effects arising from changes to human vocabulary are smaller and slower than the effects arising from direct LLM-assisted editing. Reassuringly, our estimates are in broad agreement with existing surveys of LLM usage (see above).

Note that the same limitation applies to all other approaches estimating LLM usage, including those based on frequency gaps \citep{kobak2025delving}, mixture models \citep{Liang2025}, and distribution overlaps \citep{siler2026diffusion}.

\paragraph{Policy implications}

LLMs can be a powerful tool for increasing equity by overcoming language barriers \citep{Berdejo-Espinola2023}, and indeed surveyed researchers often report using LLMs for translating manuscript drafts \citep{Kwon2025,Mohammadi2026,Wiley2025}.
It has also been argued that LLMs can make academic writing more efficient, allowing scientists to devote more time to research \citep{Filimonovic2025}. There is evidence that the widespread use of LLMs led to an increasing number of submissions and faster reviewing \citep{Qi2025}. 

On the other hand, LLM-assisted writing also carries a number of dangers. First, LLMs can hallucinate and misattribute their statements. An evidence of increased rate of hallucinated citations \citep{Walters2023,topaz2026fabricated, zhao2026llm} implies that scientists may sometimes fail to critically review LLM outputs, and suggests that LLMs can also be used for outright scientific fraud \citep{Kendall2024}. Second, the increase in writing efficiency can reduce the overall quality of published research by putting even stronger strains on the review process \citep{Gray2026} and amplifying publish-or-perish biases \citep{Kendall2024}. Third, the global dependency on LLMs can have a homogenizing effect not only on language but also on reasoning, threatening diversity as a source of innovation \citep{Sourati2026,Zhou2026_creative}. It does not help that many academic authors neglect already existing guidelines for ethical LLM usage \citep{kim2024research} and often do not disclose it \citep{Kousha2024,Kwon2025}. All these developments could reduce societal trust in the process of science as a whole \citep{Gray2026}.

Beyond that, LLM-assisted writing can carry individual dangers as well. The process of writing is closely connected to the process of thinking, and delegating writing to an LLM and only passively proofreading their outputs can lead to less reflection and sloppier arguments staying unnoticed \citep{messeri2024artificial,Tang2025}. Relying on LLMs for academic writing can also cause a loss of individual creativity \citep{Zhou2026_creative}. 

In summary, while LLM-assisted writing does have its advantages such as closing the language barrier, widespread reliance on LLMs can pose serious challenges to academia. These need to be addressed, both on an individual level by critical reflection, and on an institutional level by developing guidelines and increased safeguards against fraud and misconduct.

\section{Methods}

\paragraph{PubMed Central}

Pubmed Central (PMC) is a collection of open-access biomedical literature curated by the U.S. National Institutes of Health's (NIH) National Library of Medicine. In this work we used the PMC Open Access Subset (\url{https://pmc.ncbi.nlm.nih.gov/tools/openftlist/}),  downloading articles with licences allowing commercial as well as non-commercial use. We used the semiannual baseline published on January 23, 2026 containing papers dated up to December 31, 2025.

In our subset of 1,194,287 papers (see below), the top three journals by the number of papers were \emph{Scientific Reports} (8.1\%), \emph{Nature Communications} (4.7\%) and \emph{PLoS ONE} (3.9\%), all of which are multidisciplinary journals covering a wide range of topics in natural sciences. In our subset, 1.8\% of papers were preprints from bioRxiv, medRxiv, arXiv, and Research Square, that had funding requirement to be uploaded to PMC. If such a preprint gets published in a journal that is part of PMC, the preprint version is not replaced, so that two versions of the same paper can exist in PMC.

\paragraph{Constructing the dataset}

From the downloaded XML files, each corresponding to one paper, we extracted the article type, publication date, language, affiliation country, PMC ID, Pubmed ID, journal name, title, abstract, and also titles and texts of each section. If the language was not specified, we used Polyglot's detector module (\url{https://polyglot.readthedocs.io/}). All non-English papers were discarded.

Some publication dates only specified the month (e.g. ``Feb 2025'') or the season (e.g. ``Fall 2025''), in which case we replaced them with the first day of the month/season. All papers without date information were discarded.

For each section, we extracted its title and text, but sometimes some of the sections were untitled. If the first section was untitled or if there was some text in the body tag but outside of a section tag, we assumed it to be the Introduction. All other untitled sections were discarded. To standardize section titles, we renamed ``Experimental results'', ``Findings'', ``Main results'', etc., into ``Results'', and similarly ``Data'', ``Experimental methods'', etc., into ``Methods''. If afterwards a paper had multiple sections with the same standardized title, they were concatenated. We only kept papers containing sections titled Introduction, Methods, Results, and Discussion (this was the most common paper structure in the data). Papers with any of these sections or the abstract below 250 characters were discarded. An overview of the section lengths can be found in Table \ref{tab:section_len}.

These steps reduced the dataset from 7,125,722 to 1,602,404 papers, 1,194,287 of which were dated 2017 or later. As the affiliation country we used the country of the first affiliation of the first author, which was available for 80.2\% of the papers in our sample.

\paragraph{Standard errors}

Our estimator $(q - \hat{p}_\text{human}) / (1-\hat{p}_\text{human})$ depends on $\hat{p}_\text{human}$ and on $q$. We determined their standard errors separately and then propagated them to obtain the error for $\hat{\beta}$.

The value $q$ is a fraction of papers containing a given word (or any of the set of words), and so it has binomial variance $\operatorname{Var}[q]=q(1-q)/n$, where $n$ is the number of papers in the current month or year.

The estimate $\hat{p}_\text{human}$ is obtained by extrapolating the regression line. If $X$ is the design matrix (consisting only  of pre-ChatGPT time points on which the regression was fit and a column of ones), $X_\text{new}$ is the predictor matrix with post-ChatGPT time points, and $y$ the response vector containing pre-ChatGPT $q$ values, then 
\begin{equation*}
    \operatorname{Var}(\hat{p}_\text{human})= \hat{\sigma}^2 X_\text{new}(X^TX)^{-1}X_\text{new}^T + \hat{\sigma}^2.
\end{equation*}
Here $\hat{\sigma}^2=\|y-X\hat{b}\|^2 / (k-2)$ is the unbiased estimate of the noise variance, where $k$ is the number of time-points in the regression (60 for monthly data and 5 for yearly data), and $\hat{b}=(X^TX)^{-1}X^Ty$ are the estimated regression coefficients. The first term gives the uncertainty of the mean prediction and the second term captures expected deviations from this mean.

To compute the variance of $\hat{\beta}$, we used the delta method assuming independence of the two variance terms:
\begin{align*}
    \operatorname{Var}[\hat{\beta}] &= \frac{\partial\hat{\beta}}{\partial q} \operatorname{Var}[q] + \frac{\partial\hat{\beta}}{\partial \hat{p}_\text{human}} \operatorname{Var}[\hat p_\text{human}] \\ &= \frac{q(1-q)/n}{1-\hat{p}_\text{human}} + \frac{(q-1)\operatorname{Var}[\hat p_\text{human}]}{(1-\hat{p}_\text{human})^2}.  
\end{align*} 
When optimizing the threshold $T$, we excluded all estimates with standard error (square root of the variance) above 0.025. We also excluded all thresholds leading to $\hat{p}_\text{human} \ge 0.999$. Note that when the optimized threshold was applied to other years or affiliation countries, resulting standard errors sometimes got higher than 0.025 (Table~\ref{tab:results}).

\paragraph{Simulation experiment}

To validate our estimation procedure, we performed the same analysis on simulated data with known $p_\text{LLM}$ and $p_\text{human}$ and varying $\beta$. We simulated data for $n=100,000$ texts, which is close to the typical number of papers per year in our dataset, and for $m=500$ words, all of which were treated as LLM marker words.

For each word, its $p_\text{human}$ was sampled from a gamma distribution with shape parameter $2$ and scale parameter $0.02$. Its $p_\text{LLM}$ was set $(1+\delta)$ times larger, with $\delta \sim U(0.5,5)$. If needed, $\delta$ was repeatedly sampled until the resulting probability was below 1.

We then sampled two binary matrices, $M_\text{pre}$ and $M_\text{post}$, each with dimensions $n \times m$, corresponding to data before and after the adoption of LLMs. For $M_\text{pre}$, each entry is a Bernoulli draw with corresponding $p_\text{human}$ (corresponding to the $m$-th column). For $M_\text{post}$, $(1 - \beta) n$ rows were filled in the same way, and the remaining $\beta n$ rows used the corresponding $p_\text{LLM}$.

We then carried out our estimation procedure, treating word frequencies in $M_\text{pre}$ as $\hat{p}_\text{human}$, and those in $M_\text{post}$ as the observed frequencies $q$. All words were ordered by frequency in $M_\text{pre}$, word sets were formed for various thresholds $T$, and we found $\hat{\beta}=\max_T(\hat\beta_\text{LB})$ as the maximum over thresholds $T$, after discarding thresholds yielding $\hat{p}_\text{human}>0.999$ or standard error above 0.025. As we did not need to use regression here, the variance of $\hat{p}_\text{human}$ was computed as $\hat{p}_\text{human}(1-\hat{p}_\text{human})/n$.

\paragraph{Threshold grids}

When optimizing the threshold $T$, we used a grid  with 12 values spaced evenly on a log-scale between 0 and 1 and added several additional values in the high-interest region between 0.03 and 0.7, resulting in 19 thresholds.

When doing the analysis by country, we did not optimize the threshold $T$ for each country separately, but used a fixed set of marker words $G$ because otherwise the comparison in Figure~\ref{fig:countries}c would not be possible. To generate the fixed set of marker words, we used the full papers in our entire dataset, but employed the threshold two steps lower in our grid compared to the threshold optimized for the full dataset. This decrease was necessary to avoid unreliable estimates with large standard errors. Note that as the result, some of our estimates for individual countries should be treated as lower bounds.

\paragraph{Cropping-based analysis}

To create section segments of comparable length, every sample was cropped to a randomly placed 255-word long segment. If a sample was shorter than that, it was used as is. The length of 255 words corresponds to the median abstract length in a subset of the PMC data that we used for initial exploration and is slightly larger than the  median abstract length in the final dataset (248 words; Table~\ref{tab:section_len}).

\subsection{Data availability}

All data are openly available at the PubMed Central website (\url{https://pmc.ncbi.nlm.nih.gov/tools/openftlist/}).

\subsection{Code availability}

All code is available on Github at \url{https://github.com/kobaklab/llm-usage-in-pmc}.

\section*{Acknowledgments}

The authors would like to thank Jan Lause for discussions.

\section*{Author contributions}
All authors contributed to conceptualization and design, as well as editing the paper. L.H. also performed all statistical analyses and drafted the paper.

\section*{Competing Interests}
The authors declare no competing interests.
\bibliographystyle{plainnat}
\bibliography{references.bib}

@article{zhao2026llm,
  title={{LLM} hallucinations in the wild: Large-scale evidence from non-existent citations},
  author={Zhao, Zhenyue and Wang, Yihe and Stuart, Toby and De Vaan, Mathijs and Ginsparg, Paul and Yin, Yian},
  journal={arXiv preprint arXiv:2605.07723},
  year={2026}
}

@article{siler2026diffusion,
  title={The diffusion of large language models in published academic articles},
  author={Siler, Kyle},
  journal={Proceedings of the National Academy of Sciences},
  volume={123},
  number={22},
  pages={e2605754123},
  year={2026},
  publisher={National Academy of Sciences}
}

@article{kobak2025delving,
  title={Delving into {LLM}-assisted writing in biomedical publications through excess vocabulary},
  author={Kobak, Dmitry and Gonz{\'a}lez-M{\'a}rquez, Rita and Horv{\'a}t, Em{\H{o}}ke-{\'A}gnes and Lause, Jan},
  journal={Science Advances},
  volume={11},
  number={27},
  pages={eadt3813},
  year={2025},
  publisher={American Association for the Advancement of Science}
}

@article{gray2024chatgpt,
  title={Chat{GPT} ``contamination'': estimating the prevalence of {LLM}s in the scholarly literature},
  author={Gray, Andrew},
  journal={arXiv preprint arXiv:2403.16887},
  year={2024}
}

@article{topaz2026fabricated,
  title={Fabricated citations: an audit across 2{\textperiodcentered} 5 million biomedical papers},
  author={Topaz, Maxim and Roguin, Nir and Gupta, Pallavi and Zhang, Zhihong and Peltonen, Laura-Maria},
  journal={The Lancet},
  volume={407},
  number={10541},
  pages={1779--1781},
  year={2026},
  publisher={Elsevier}
}

@misc{Wiley2025,
   author = {Wiley},
   title = {Explan{AI}tions: An {AI} study by Wiley},
   url = {https://www.wiley.com/en-de/about-us/ai-resources/ai-study/},
   month = {2},
   year = {2025}
}

@misc{Wiley2026,
   author = {Wiley},
   title = {Explan{AI}tions 2025-2026: The evolution of {AI} in research.},
   url = {https://www.wiley.com/en-de/about-us/ai-resources/ai-study/},
   month = {2},
   year = {2026}
}

@misc{pmc,
  author       = {National Library of Medicine},
  title        = {PMC Open Access Subset [Internet]},
  year         = {2003},
  address      = {Bethesda (MD)},
  howpublished = {\url{https://pmc.ncbi.nlm.nih.gov/tools/openftlist/}},
  note         = {[cited 2026 07 04]}
}

@article{messeri2024artificial,
  title={Artificial intelligence and illusions of understanding in scientific research},
  author={Messeri, Lisa and Crockett, Molly J},
  journal={Nature},
  volume={627},
  number={8002},
  pages={49--58},
  year={2024},
  publisher={Nature Publishing Group UK London}
}

@article{kim2024research,
  title={Research ethics and issues regarding the use of {C}hat{GPT}-like artificial intelligence platforms by authors and reviewers: a narrative review},
  author={Kim, Sang-Jun},
  journal={Science Editing},
  volume={11},
  number={2},
  pages={96--106},
  year={2024},
  publisher={Korean Council of Science Editors}
}

@article{Akram2024,
  title={Quantitative analysis of {AI}-generated texts in academic research: A study of {AI} presence in Arxiv submissions using {AI} detection tool},
  author={Akram, Arslan},
  journal={arXiv preprint arXiv:2403.13812},
  year={2024}
}

@article{Astarita2024,
  title={Delving into the utilisation of {C}hat{GPT} in scientific publications in astronomy},
  author={Astarita, Simone and Kruk, Sandor and Reerink, Jan and G{\'o}mez, Pablo},
  journal={arXiv preprint arXiv:2406.17324},
  year={2024}
}

@article{Bao2025_classifier,
   author = {Honglin Bao and Mengyi Sun and Misha Teplitskiy},
   journal = {Quantitative Science Studies},
   month = {8},
   pages = {716-731},
   publisher = {MIT Press},
   title = {Where there’s a will there’s a way: {C}hat{GPT} is used more for science in countries where it is prohibited},
   volume = {6},
   year = {2025}
}

@article{Berdejo-Espinola2023,
   author = {Violeta Berdejo-Espinola and Tatsuya Amano},
   journal = {Science},
   month = {3},
   pages = {991},
   pmid = {36893248},
   publisher = {American Association for the Advancement of Science},
   title = {{AI} tools can improve equity in science},
   volume = {379},
   year = {2023}
}

@misc{Botes2025,
  title={Initial indications of generative {AI} writing in linguistics research publications},
  author={Botes, Elouise and Dewaele, Jean-Marc and Colling, Joanne and Teuber, Ziwen},
  year={2025},
  publisher={PsyArXiv. https://doi. org/10.31234/osf. io/4yvbp\_v1}
}

@inproceedings{Dou2024,
  title={Enhancing robustness of {LLM}-synthetic text detectors for academic writing: A comprehensive analysis},
  author={Dou, Zhicheng and Guo, Yuchen and Chang, Ching-Chun and Nguyen, Huy H and Echizen, Isao},
  booktitle={International Conference on Advanced Information Networking and Applications},
  pages={266--277},
  year={2024},
  organization={Springer}
}

@misc{Elsevier2024,
   author = {Elsevier},
   title = {Insights 2024 | Attitudes toward {AI} | {E}lsevier},
   url = {https://web.archive.org/web/20260410222804/https://www.elsevier.com/insights/attitudes-toward-ai/the-current-ai-landscape},
   year = {2024}
}

@article{Eppler2024,
   author = {Michael Eppler and Conner Ganjavi and Lorenzo Storino Ramacciotti and Pietro Piazza and Severin Rodler and Enrico Checcucci and Juan Gomez Rivas and Karl F. Kowalewski and Ines Rivero Belenchón and Stefano Puliatti and Mark Taratkin and Alessandro Veccia and Loïc Baekelandt and Jeremy Y.C. Teoh and Bhaskar K. Somani and Marcelo Wroclawski and Andre Abreu and Francesco Porpiglia and Inderbir S. Gill and Declan G. Murphy and David Canes and Giovanni E. Cacciamani},
   journal = {European Urology},
   month = {2},
   pages = {146-153},
   pmid = {38182492},
   publisher = {Elsevier},
   title = {Awareness and Use of {C}hat{GPT} and {L}arge {L}anguage {M}odels: A Prospective Cross-sectional Global Survey in Urology},
   volume = {85},
   year = {2024}
}

@article{Filimonovic2025,
  title={Can {G}en{AI} improve academic performance? Evidence from the social and behavioral sciences},
  author={Filimonovic, Dragan and Rutzer, Christian and Wunsch, Conny},
  journal={arXiv preprint arXiv:2510.02408},
  year={2025}
}

@inproceedings{Geng2025,
   author = {Mingmeng Geng and Roberto Trotta},
   booktitle={Findings of the Association for Computational Linguistics: ACL 2025},
   month = {2},
   pages = {12689-12696},
   title = {Human-{LLM} Coevolution: Evidence from Academic Writing},
   year = {2025}
}

@article{Geng2024,
  title={Is {C}hat{GPT} Transforming Academics' Writing Style?},
  author={Geng, Mingmeng and Trotta, Roberto},
  journal={arXiv preprint arXiv:2404.08627},
  year={2024}
}

@article{Gray2026,
   author = {Andrew Gray},
   journal = {Taylor \& Francis},
   pages = {85-88},
   publisher = {Taylor and Francis Ltd.},
   title = {How {AI} use in scholarly publishing threatens research integrity, lessens trust, and invites misinformation},
   volume = {82},
   year = {2026}
}

@article{Gray2025,
  title={Estimating the prevalence of {LLM}-assisted text in scholarly writing},
  author={Gray, Andrew},
  journal={arXiv preprint arXiv:2512.01560},
  year={2025}
}

@article{Kendall2024,
   author = {Graham Kendall and Jaime A Teixeira Da Silva},
   journal = {Learned Publishing},
   title = {Risks of abuse of large language models, like {C}hat{GPT}, in scientific publishing: Authorship, predatory publishing, and paper mills},
   volume = {37},
   year = {2024}
}

@article{Kousha2025,
   author = {Kayvan Kousha and Mike Thelwall},
   journal = {Scientometrics 2026},
   month = {9},
   pages = {1-21},
   publisher = {Springer},
   title = {How much are {LLM}s changing the language of academic papers after {C}hat{GPT}? A multi-database and full text analysis},
   year = {2025}
}

@article{Kousha2024,
   author = {Kayvan Kousha},
   journal = {Scientometrics},
   month = {12},
   pages = {7959-7969},
   publisher = {Akademiai Kiado ZRt.},
   title = {How is {C}hat{GPT} acknowledged in academic publications?},
   volume = {129},
   year = {2024}
}

@article{Kwon2025,
   author = {Diana Kwon},
   journal = {Nature},
   month = {5},
   pages = {574-578},
   pmid = {40369145},
   publisher = {Nature Research},
   title = {Is it {OK} for {AI} to write science papers? Nature survey shows researchers are split},
   volume = {641},
   year = {2025}
}

@article{Lazebnik2024,
   author = {Teddy Lazebnik and Ariel Rosenfeld},
   journal = {JDIS Journal of Data and Information Science},
   title = {Detecting {LLM}-assisted writing in scientific communication: Are we there yet?},
   year = {2024}
}

@inproceedings{liang2024monitoring,
  title={Monitoring {AI}-modified content at scale: a case study on the impact of {ChatGPT} on {AI} conference peer reviews},
  author={Liang, Weixin and Izzo, Zachary and Zhang, Yaohui and Lepp, Haley and Cao, Hancheng and Zhao, Xuandong and Chen, Lingjiao and Ye, Haotian and Liu, Sheng and Huang, Zhi and others},
  booktitle={Proceedings of the 41st International Conference on Machine Learning},
  pages={29575--29620},
  year={2024}
}

@article{Liang2024_paper,
  title={Mapping the increasing use of {LLM}s in scientific papers},
  author={Liang, Weixin and Zhang, Yaohui and Wu, Zhengxuan and Lepp, Haley and Ji, Wenlong and Zhao, Xuandong and Cao, Hancheng and Liu, Sheng and He, Siyu and Huang, Zhi and others},
  journal={arXiv preprint arXiv:2404.01268},
  year={2024}
}

@article{Liang2025,
   author = {Weixin Liang and Yaohui Zhang and Zhengxuan Wu and Haley Lepp and Wenlong Ji and Xuandong Zhao and Hancheng Cao and Sheng Liu and Siyu He and Zhi Huang and Diyi Yang and Christopher Potts and Christopher D. Manning and James Zou},
   journal = {Nature Human Behaviour},
   month = {8},
   pages = {2599-2609},
   pmid = {40760036},
   publisher = {Nature Publishing Group},
   title = {Quantifying large language model usage in scientific papers},
   volume = {9},
   year = {2025}
}

@article{Liao2024,
  title={Llms as research tools: A large scale survey of researchers' usage and perceptions},
  author={Liao, Zhehui and Antoniak, Maria and Cheong, Inyoung and Cheng, Evie Yu-Yen and Lee, Ai-Heng and Lo, Kyle and Chang, Joseph Chee and Zhang, Amy X},
  journal={arXiv preprint arXiv:2411.05025},
  year={2024}
}

@article{Lin2025_NES,
  title={{C}hat{GPT} as linguistic equalizer? quantifying {LLM}-Driven lexical shifts in academic writing},
  author={Lin, Dingkang and Zhao, Naixuan and Tian, Dan and Li, Jiang},
  journal={arXiv preprint arXiv:2504.12317},
  year={2025}
}

@article{Liu2024_profiles,
  title={Towards the relationship between {AIGC} in manuscript writing and author profiles: evidence from preprints in {LLM}s},
  author={Liu, Jialin and Bu, Yi},
  journal={arXiv preprint arXiv:2404.15799},
  year={2024}
}

@article{Matsui2024,
  title={Delving into {P}ub{M}ed records: Some terms in medical writing have drastically changed after the arrival of {C}hat{GPT}},
  author={Matsui, Kentaro},
  journal={MedRxiv},
  pages={2024--05},
  year={2024},
  publisher={Cold Spring Harbor Laboratory Press}
}

@article{Mishra2024,
   author = {Tanisha Mishra and Edward Sutanto and Rini Rossanti and Nayana Pant and Anum Ashraf and Akshay Raut and Germaine Uwabareze and Ajayi Oluwatomiwa and Bushra Zeeshan},
   journal = {Scientific Reports},
   pages = {31672},
   title = {Use of large language models as artificial intelligence tools in academic research and publishing among global clinical researchers},
   volume = {14},
   year = {2024}
}

@article{Mohammadi2026,
   author = {Ehsan Mohammadi and Mike Thelwall and Yizhou Cai and Taylor Collier and Iman Tahamtan and Azar Eftekhar},
   journal = {Information Processing \& Management},
   month = {1},
   pages = {104350},
   publisher = {Pergamon},
   title = {Is generative {AI} reshaping academic practices worldwide? A survey of adoption, benefits, and concerns},
   volume = {63},
   year = {2026}
}

@article{Ng2025,
   author = {Jeremy Y. Ng and Sharleen G. Maduranayagam and Nirekah Suthakar and Amy Li and Cynthia Lokker and Alfonso Iorio and R. Brian Haynes and David Moher},
   journal = {The Lancet Digital Health},
   month = {1},
   pages = {e94-e102},
   pmid = {39550312},
   publisher = {Elsevier},
   title = {Attitudes and perceptions of medical researchers towards the use of artificial intelligence chatbots in the scientific process: an international cross-sectional survey},
   volume = {7},
   year = {2025}
}

@article{Nordling2023,
   author = {Linda Nordling},
   journal = {Nature},
   month = {10},
   pages = {655-657},
   pmid = {37845528},
   publisher = {Nature Research},
   title = {How {C}hat{GPT} is transforming the postdoc experience},
   volume = {622},
   year = {2023}
}

@article{Picazo-Sanchez2024,
   author = {Pablo Picazo-Sanchez and Lara Ortiz-Martin},
   journal = {Applied Intelligence},
   month = {3},
   pages = {4172-4188},
   publisher = {Springer},
   title = {Analysing the impact of {C}hat{GPT} in research},
   volume = {54},
   year = {2024}
}

@article{Prakash2025,
   author = {Arjun Prakash and Shruti Aggarwal and Jeevan John Varghese and Joel John Varghese},
   journal = {Humanities and Social Sciences Communications 2025 12:1},
   month = {7},
   pages = {1058-},
   publisher = {Palgrave},
   title = {Writing without borders: {AI} and cross-cultural convergence in academic writing quality},
   volume = {12},
   year = {2025}
}

@article{Qi2025,
  title={Does GenAI Rewrite How We Write? An Empirical Study on Two-Million Preprints},
  author={Qi, Minfeng and Cao, Zhongmin and Wang, Qin and Li, Ningran and Zhu, Tianqing},
  journal={arXiv preprint arXiv:2510.17882},
  year={2025}
}

@article{Sourati2026,
   author = {Zhivar Sourati and Alireza S. Ziabari and Morteza Dehghani},
   journal = {Trends in Cognitive Sciences},
   month = {1},
   publisher = {Elsevier},
   title = {The Homogenizing Effect of Large Language Models on Human Expression and Thought},
   year = {2026}
}

@article{Tang2025,
   author = {Bor Luen Tang},
   journal = {Publications 2025},
   month = {12},
   pages = {63},
   publisher = {Multidisciplinary Digital Publishing Institute},
   title = {The Epistemic Downside of Using {LLM}-Based Generative {AI} in Academic Writing},
   volume = {13},
   year = {2025}
}

@article{Uribe2024,
   author = {Sergio E. Uribe and Ilze Maldupa},
   journal = {Journal of Dentistry},
   month = {10},
   pages = {105275},
   pmid = {39089668},
   publisher = {Elsevier},
   title = {Estimating the use of {C}hat{GPT} in dental research publications},
   volume = {149},
   year = {2024}
}

@article{VanNoorden2023,
   author = {Richard Van Noorden and Jeffrey M. Perkel},
   doi = {10.1038/D41586-023-02980-0},
   issn = {14764687},
   issue = {7980},
   journal = {Nature},
   month = {9},
   pages = {672-675},
   pmid = {37758894},
   publisher = {Nature Research},
   title = {{AI} and science: what 1,600 researchers think},
   volume = {621},
   year = {2023}
}

@article{Walters2023,
   author = {William H. Walters and Esther Isabelle Wilder},
   journal = {Scientific Reports},
   month = {9},
   pages = {14045-},
   pmid = {37679503},
   publisher = {Nature Publishing Group},
   title = {Fabrication and errors in the bibliographic citations generated by {C}hat{GPT}},
   volume = {13},
   year = {2023}
}

@article{Weber-Wulff2023,
   author = {Debora Weber-Wulff and Alla Anohina-Naumeca and Sonja Bjelobaba and Tomáš Foltýnek and Jean Guerrero-Dib and Olumide Popoola and Petr Šigut and Lorna Waddington},
   journal = {International Journal for Educational Integrity},
   month = {12},
   publisher = {BioMed Central Ltd},
   title = {Testing of detection tools for {AI}-generated text},
   volume = {19},
   year = {2023}
}

@article{Yakura2025,
  title={Empirical evidence of Large Language Model's influence on human spoken communication},
  author={Yakura, Hiromu and Lopez-Lopez, Ezequiel and Brinkmann, Levin and Serna, Ignacio and Gupta, Prateek and Soraperra, Ivan and Rahwan, Iyad},
  journal={arXiv preprint arXiv:2409.01754},
  year={2024}
}

@article{Zhou2026_creative,
   author = {Yiyong Zhou and Qinghan Liu and Jihao Huang and Guiquan Li},
   journal = {Technology in Society},
   month = {3},
   pages = {103087},
   publisher = {Pergamon},
   title = {Creative scar without generative {AI}: Individual creativity fails to sustain while homogeneity keeps climbing},
   volume = {84},
   year = {2026}
}

\clearpage

\onecolumn

\section*{Supplementary Tables}
\renewcommand{\thefigure}{S\arabic{figure}}
\setcounter{figure}{0}  
\renewcommand{\thetable}{S\arabic{table}}
\setcounter{table}{0} 

\begin{table}[h]
\centering
\begin{tabular}{cllcll}
\toprule
\textbf{Period} & \textbf{Section} & \textbf{Dataset} & \textbf{Estimate} & \textbf{Method} & \textbf{Ref.} \\
\midrule
12/2022--02/2023 & abstract & Various journals & 0.10 & Various LLM detectors & \citeyear{Picazo-Sanchez2024}\\
\midrule
08/2023 & abstract  & arXiv, bioRxiv    & 0.13 & LLM detector (custom)  & \citeyear{Bao2025_classifier}\\
\midrule
11/2023 & full paper      & arXiv (CS)        & 0.07 & LLM detector (Originality.AI) & \citeyear{Akram2024}\\
\midrule
2023    & full paper      & Dimensions        & 0.02 & frequency gap     & \citeyear{gray2024chatgpt}\\
\midrule
\multirow{2}{*}{2023}    & \multirow{2}{*}{abstract}  & \multirow{2}{*}{arXiv NLP papers}    & 0.07 & LLM detector (GPTKIT)& \multirow{2}{*}{\citeyear{Liu2024_profiles}} \\
        &           &                   & 0.04 & LLM detector (Smodin) & \\
\midrule
01/2024 & abstract  & arXiv (CS)        & 0.35 & mixture model & \citeyear{Geng2024}\\
\midrule
\multirow{4}{*}{02/2024} & abstract  & arXiv (CS)        & 0.18 & \multirow{4}{*}{mixture model} & \multirow{4}{*}{\citeyear{Liang2024_paper}}\\
        & introduction & arxiv (CS)          & 0.14 & & \\
        & abstract  & Nature portfolio  & 0.06 & & \\
        & introduction & Nature portfolio  & 0.04 & & \\
\midrule
\multirow{4}{*}{09/2024} & abstract  & arXiv (CS)        & 0.23 & \multirow{4}{*}{mixture model} & \multirow{4}{*}{\citeyear{Liang2025}}\\
        & introduction & arxiv (CS)         & 0.20 & & \\
        & abstract  & Nature portfolio  & 0.09 & & \\
        & introduction & Nature portfolio  & 0.09 & & \\
\midrule
2024    & abstract  & PubMed            & 0.14 & frequency gap     & \citeyear{kobak2025delving}\\
\midrule
2024    & full paper      & Dimensions        & 0.12 & frequency gap    & \citeyear{Gray2025}\\
\midrule
2024    & full paper     & Dim., OpenAlex, PMC & 0.16 & frequency gap    & \citeyear{Kousha2025}\\
\midrule
2023    & \multirow{3}{*}{full paper}    & \multirow{3}{*}{\makecell[l]{Elsevier, MDPI, \\Frontiers, PLoS}} & 0.12 & \multirow{3}{*}{distribution overlap}&\multirow{3}{*}{\citeyear{siler2026diffusion}}\\
2024    &&& 0.36 &\\
2025    &&& 0.57 &\\

\bottomrule
\end{tabular}
\caption{\textbf{Literature overview: Estimates of LLM usage in academia.} Dimensions and OpenAlex are databases containing academic papers from diverse disciplines. All studies on LLM detectors used existing detectors, except for \citet{Bao2025_classifier}, who trained their own. We excluded one of the detectors used by \citet{Liu2024_profiles} (Sapling) because it returned high LLM usage for pre-2022 papers.}
\label{tab:review_estimation}
\end{table}

\begin{table}
\centering
\begin{tabular}{ccllcl}
\toprule
\textbf{Period} & \textbf{Sample size} & \textbf{Sample} & \textbf{LLMs used for} & \textbf{Usage} & \textbf{Ref.} \\
\midrule

04--05/2023
& 456
& urologists
& Summarizing text & 27\% 
& \citeyear{Eppler2024}\\

\midrule

06--07/2023
& 3,838
& postdocs 
&  \makecell[l]{Refining text} 
& 20\% 
& \citeyear{Nordling2023}\\

\midrule

07--08/2023
& 2,165
& \makecell[l]{PubMed authors} 
& \makecell[l]{Writing or editing manuscripts} 
& 22\% 
& \citeyear{Ng2025}\\

\midrule

?--09/2023
& 1,659 
& \makecell[l]{recent authors} 
& \makecell[l]{To help write research manuscripts} 
& 21\% 
& \citeyear{VanNoorden2023}\\

\midrule

\multirow{5}{*}{11/2023--04/2024} 
& \multirow{5}{*}{816} 
& \multirow{5}{*}{\makecell[l]{researchers\\(Semantic Scholar)}} 
& Fix grammar or rephrase 
& 50\% 
& \multirow{5}{*}{\citeyear{Liao2024}}\\
&&& Rewrite for another style & 24\% &\\
&&& Shorten or summarize & 23\% &\\
&&& Draft paragraphs from ideas & 21\% &\\
&&& Direct writing (overall) & 43\% &\\

\midrule

\multirow{3}{*}{01--03/2024}
& \multirow{3}{*}{2,021} 
& \multirow{3}{*}{\makecell[l]{researchers\\(Scopus)}} 
& Synthesizing drafts 
& 16\%
& \multirow{3}{*}{\citeyear{Mohammadi2026}}\\
&&& Proofreading drafts & 17\% &\\
&&& Translating text & 18\% &\\

\midrule

\multirow{2}{*}{03--04/2024} 
& \multirow{2}{*}{1,043} 
& \multirow{2}{*}{researchers} 
& Help with translation 
& 40\% 
& \multirow{2}{*}{\citeyear{Wiley2025}}\\
&&& Proofreading and editing & 38\% &\\

\midrule

\multirow{4}{*}{03--05/2024}
& \multirow{4}{*}{5,229} 
& \multirow{4}{*}{researchers} 
& To edit your paper 
& 28\% 
& \multirow{4}{*}{\citeyear{Kwon2025}}\\
&&& To translate your paper & 8\% &\\
&&& \makecell[l]{To summarize other papers} 
& 8\% &\\
&&& \makecell[l]{To write the first draft} & 8\% &\\

\midrule

\multirow{2}{*}{04--06/2024} 
& \multirow{2}{*}{226} 
& \multirow{2}{*}{\makecell[l]{medical researchers}} 
& Writing 
& 8\% 
& \multirow{2}{*}{\citeyear{Mishra2024}}\\
&&& Revision and editing & 8\% &\\

\midrule

07--08/2025
& 341--557 
& academia, healthcare, etc.
& \makecell[l]{Writing assistance\\(copyediting, translation, etc.)} 
& 71\%
& \citeyear{Wiley2026}\\

\bottomrule
\end{tabular}
\caption{\textbf{Literature overview: Surveys on LLM usage in academia.} We only included surveys that explicitly mention LLM use-cases that are likely to have an effect on the published manuscript text, such as translation, editing, etc. We excluded questions about LLM usage in general, or about use-cases with limited influence on the text, such as grammar checks. The use-cases are copied verbatim from the surveys, and sometimes abbreviated.}
\label{tab:review_surveys}
\end{table}

\begin{table}[]
    \centering
    \begin{tabular}{lcccr}
    \toprule
    \textbf{Country} & \textbf{2022 frequency} & \textbf{2025 frequency} & \textbf{LLM usage} & \textbf{Sample size} \\
    \midrule
    Entire PMC corpus       & 0.83 & 0.95 & 0.68 \scriptsize{(2)} & 1,194,287 \\
    \midrule
    Majority native English & 0.92 & 0.96 & 0.37 \scriptsize{(3)} & 333,161 \\
    Other countries         & 0.79 & 0.95 & 0.72 \scriptsize{(2)} & 719,959 \\
    \midrule
    China                   & 0.75 & 0.97 & 0.82 \scriptsize{(2)} & 246,000 \\
    USA                     & 0.93 & 0.97 & 0.39 \scriptsize{(5)} & 208,178 \\
    UK                      & 0.91 & 0.95 & 0.28 \scriptsize{(7)} & 64,347 \\
    Japan                   & 0.68 & 0.86 & 0.55 \scriptsize{(4)} & 59,863 \\
    Germany                 & 0.90 & 0.96 & 0.53 \scriptsize{(4)} & 54,827 \\
    Italy                   & 0.85 & 0.95 & 0.61 \scriptsize{(5)} & 28,716 \\
    Canada                  & 0.92 & 0.96 & 0.46 \scriptsize{(14)}& 28,649 \\
    Netherlands             & 0.86 & 0.95 & 0.49 \scriptsize{(5)} & 27,538 \\
    South Korea             & 0.72 & 0.96 & 0.85 \scriptsize{(2)} & 25,537 \\
    France                  & 0.88 & 0.93 & 0.33 \scriptsize{(11)}& 24,577 \\
    Australia               & 0.91 & 0.95 & 0.31 \scriptsize{(11)}& 22,956 \\
    Spain                   & 0.86 & 0.97 & 0.70 \scriptsize{(7)} & 22,657 \\
    Brazil                  & 0.82 & 0.94 & 0.63 \scriptsize{(4)} & 19,670 \\
    Sweden                  & 0.87 & 0.93 & 0.47 \scriptsize{(11)}& 16,015 \\
    Switzerland             & 0.90 & 0.97 & 0.67 \scriptsize{(7)} & 15,389 \\
    India                   & 0.86 & 0.96 & 0.69 \scriptsize{(5)} & 14,173 \\
    Taiwan                  & 0.77 & 0.96 & 0.80 \scriptsize{(4)} & 11,001 \\
    Turkey                 & 0.64 & 0.90 & 0.67 \scriptsize{(5)} & 10,287 \\
    Iran                    & 0.68 & 0.94 & 0.78 \scriptsize{(3)} & 9,399 \\
    Poland                  & 0.83 & 0.94 & 0.48 \scriptsize{(8)} & 8,210 \\
    \bottomrule
    \end{tabular}
    \caption{\textbf{Estimated LLM usage in 2025 for individual countries (full paper).} Top 20 countries by the number of papers are sorted by sample size. The results for countries with majority native English speakers include all PMC countries with a majority of native English speakers, not only those from the top 20. The reported frequencies correspond to the fixed set of marker words identified using the entire PMC corpus, but using a lower threshold value compared to what we used in Table~\ref{tab:results} (see Methods). This explains the difference between the first row in this table and Table~\ref{tab:results}.}
    \label{tab:countries}
\end{table}

\begin{table}[]
    \centering
    \begin{tabular}{lccc}
    \toprule
    \textbf{Section} & \textbf{5th percentile} & \textbf{Median} & \textbf{95th percentile} \\
    \midrule
    Abstract & 148 & 248 & 449 \\
    Introduction & 246 & 595 & 1,314 \\
    Methods & 433 & 1,270 & 3,452 \\
    Results & 487 & 1,852 & 5,354 \\
    Discussion & 510 & 1,179 & 2,284 \\
    Full paper & 2,527 & 5,156 & 10,684 \\
    \bottomrule
    \end{tabular}
    \caption{\textbf{Section lengths in words.}}
    \label{tab:section_len}
\end{table}

\end{document}